%% file: root.tex
\documentclass[letterpaper, 10 pt, conference]{ieeeconf}  %

\usepackage{graphics} %
\usepackage{times} %
\usepackage{amsmath} %
\usepackage{amssymb}  %
\usepackage{todonotes}
\usepackage{booktabs}
\usepackage{hyperref}
\usepackage{cleveref}
\usepackage{subcaption}
\usepackage{bm}
\usepackage{soul}
\usepackage{multirow}
\usepackage{float}
\usepackage{tabularx}
\usepackage{gensymb}

\usepackage{algorithm}
\usepackage[noend]{algpseudocode}

\newcommand{\proc}[1]{\textsc{#1}}
\algnewcommand\ENDPROCEDURE{\item[\algorithmicendprocedure]}%

\input{inc-macros}

\title{Combining Planning and Diffusion for Mobility with Unknown Dynamics}

\author{Yajvan M Ravan
\quad Zhutian Yang\quad Tao Chen\quad 
Tomás Lozano-Pérez\quad
Leslie Pack Kaelbling\\
Massachusetts Institute of Technology
}

\begin{document}

\maketitle
\thispagestyle{empty}
\pagestyle{empty}

\input{text/1-abstract}

\input{text/2-intro-new}
\input{text/3-related-new}

\input{text/4-method-new}

\input{text/5-experiments-new}
\input{text/6-conclusion}
\input{text/7-acknowledgements}

{\small
\bibliographystyle{IEEEtran}
\bibliography{references}
}

\end{document}

%% file: inc-macros.tex
\usepackage{xspace}

\definecolor{MyRed}{rgb}{1.0,0.0,0.0}
\definecolor{MyBlue}{rgb}{0.204,0.451,0.859}

\newcommand{\oursfull}{Planner-Ordered Policy\xspace}
\newcommand{\ours}{\textit{PoPi}\xspace}

%% file: text/1-abstract.tex
\begin{abstract} 
Manipulation of large objects over long horizons (such as carts in a warehouse) is an essential skill for deployable robotic systems. Large objects require mobile manipulation which involves simultaneous manipulation, navigation, and movement with the object in tow. In many real-world situations, object dynamics are incredibly complex, such as the interaction of an office chair (with a rotating base and five caster wheels) and the ground. We present a hierarchical algorithm for long-horizon robot manipulation problems in which the dynamics are partially unknown. We observe that diffusion-based behavior cloning is highly effective for short-horizon problems with unknown dynamics, so we decompose the problem into an abstract high-level, obstacle-aware motion-planning problem that produces a waypoint sequence. We use a short-horizon, relative-motion diffusion policy to achieve the waypoints in sequence. We train mobile manipulation policies on a Spot robot that has to push and pull an office chair. Our hierarchical manipulation policy performs consistently better, especially when the horizon increases, compared to a diffusion policy trained on long-horizon demonstrations or motion planning assuming a rigidly-attached object (success rate of 8 (versus 0 and 5 respectively) out of 10 runs). Importantly, our learned policy generalizes to new layouts, grasps, chairs, and flooring that induces more friction, without any further training, showing promise for other complex mobile manipulation problems. Project Page: \url{https://yravan.github.io/plannerorderedpolicy/}
\end{abstract}
\vspace{-5pt}

%% file: text/2-intro-new.tex
\section{Introduction}
Many robot tasks involve finding and following a path while interacting with an environment whose dynamics are not known.  For example, a robot arm pushing an object among obstacles on a table or a mobile robot pushing an office chair among furniture are both facing this type of problem.  In this paper, we explore in detail the problem of rearranging large objects (comparable to robot size) through pushing and pulling.

\begin{figure}[h]
    \centering
    \includegraphics[width=\linewidth]{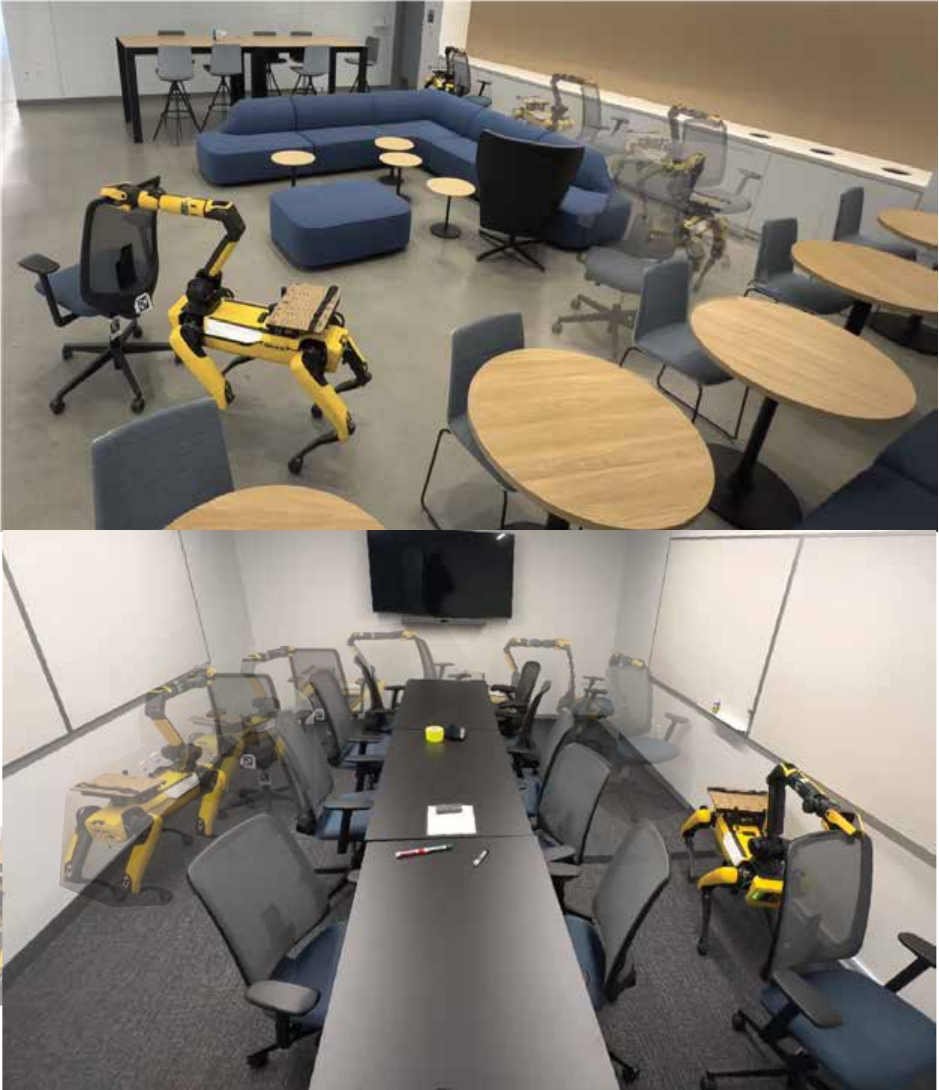}
    \caption{\small The Spot robot moving a chair to target location while, navigating among obstacles. The top environment is where training demonstrations were collected. Our hierarchical policy \ours achieves 80-100\% success rate on tests in this environment. The bottom is an unseen testing environment with higher-friction carpet \& narrower pathways. \ours generalizes zero-shot with 70\% success.}
    \label{fig:spot-chair}
\vspace{-15pt}
\end{figure}

We focus in detail on the problem of having a Boston Dynamics Spot pull a 5-wheeled office chair among other furniture (see \cref{fig:spot-chair}).  This is a challenging instance of finding and following a path subject to unknown dynamics, as the surface of the floor may be variable and may have variable friction.  Note that the effect of pushing or pulling on the chair depends on the (unobservable) orientations of the 5 casters on the legs.  Also, the robot is holding the top of the chair, which can rotate and incline.  The most common failure modes are the robot losing its grasp when making sharp turns around obstacles, which involve substantial re-orientation of the casters, or the chair colliding with/getting stuck on another piece of furniture.

We seek an approach that (a) allows the robot to be trained quickly in the real world, without access to a simulator as the complex dynamics present a large sim-to-real gap, and (b) generalizes to somewhat different environments, e.g. a different room with a carpet or a different chair.  We make two simplifying assumptions.  First is that a map of the obstacles is available. This can be easily obtained with a simple RGB-D scan of the room or existing SLAM algorithms. Second is that the environment dynamics remains roughly unchanging along the paths to be followed.  Our approach is hierarchical: we learn a ``local'' motion control policy via imitation learning and use a ``global'' motion planner to define waypoints for the local policy.  Lastly, we assume that the details of the local policy will not affect the choice of waypoints.  We call our approach \oursfull, \ours for short.

At the low-level, we learn a short-horizon, diffusion-based manipulation policy that is conditioned on pose estimates of the chair and predicts desired robot motions for reaching a waypoint. We chose this approach for its ability to efficiently learn policies from relatively few suboptimal demonstrations.  Instead of using long-horizon demonstrations as imitation learning episodes, we sample snippets from the demonstrations to learn how to perform relative changes to the object pose. This short-horizon imitation learning limits the action and observation space, thus narrowing the distribution that our low-level control policy must learn.   We then combine this policy with a high-level motion planning algorithm that uses a map generated from partial point-clouds for navigation. Thus, our imitation-learned motion-control policy need not learn how to interact with obstacles or specific environments, as this is done by motion planning.  This hierarchical approach also enables substantial generalization to new environments.

We evaluate \ours  on the task of moving office chairs among stationary furniture. We collected real-world demonstrations in one environment (35 episodes in approximately 60 minutes) and evaluate in that environment for different start and goal locations, different grasps, and different chairs. We also evaluate in a carpeted environment with substantially different dynamics. Compared to a ``global'' diffusion policy trained on long-horizon demonstrations and also compared with pure motion planning (assuming the chair remains fixed with respect to the robot), our hierarchical strategy improves long-horizon success and generalizes better across objects, initial conditions, and environments.

%% file: text/3-related-new.tex
\section{Related Work}
The class of problems of interest in this paper (manipulating objects with unknown dynamics) have been investigated using a wide variety of methods, which we categorize as follows:
(1) using a simple {\it a priori} model of the dynamics and applying feedback control;
(2) motion planning from an approximate or learned model;
(3) learning a policy via reinforcement learning; and 
(4) learning a policy via imitation learning.
A review of the general area of pushing manipulation is available~\cite{frobt.2020.00008}.  Below, we highlight some of the most relevant work.

\textbf{Feedback} Heins {\it et al.}~\cite{diffikmobilemanip} present an example of a mobile manipulation system using motion planning (differential IK controller) for pushing objects and obstacle avoidance in unknown environments.  It leverages a simple model of pushing and very fast kinematic control.  This approach would be applicable to our problem, but it requires substantial  engineering effort to apply it to a new situation.

\textbf{Motion planning} There are quite a few motion planning approaches that exploit analytic or learned models of the dynamics.  The closest to our work is Zito {\it et al.}~\cite{Zito2012TwolevelRP}, who develop a two-level RRT-based push planner which, like our method, uses a high-level planner to generate sub-goals for a lower-level planner. However, their lower level uses a pushing simulator to implement a kinodynamic RRT to reach the sub-goals.  In general, the motion planning methods require a reasonably accurate model of the dynamics.  The dynamics of our problem are quite complex and difficult to model, particularly since we cannot observe the state of the wheels.

\textbf{Reinforcement learning} The use of hierarchical policies for mobile manipulation is popular. Many prior works use reinforcement learning to acquire low-level policies for primitive skills, tracking end-effector pose, or tracking whole body velocities and combine these with high-level policies that output trajectories of the aforementioned primitives \cite{visualquadrupedloco, wholebodyquadruped, gu2022multi, umionlegs, member, ma2022combining}. Chappellet {\it et al.}~\cite{humanoidloco} use 3D visual tracking along with SLAM to execute primitive actions for manipulating large objects such as wheelbarrows and bobbins instead, while Tang {\it et al.}~\cite{mobilepushing} use nonlinear model-predictive control to achieve stable pushing, without grasping, for nonholonomic robots.  Several others~\cite{visualquadrupedloco,mobilealoha, umionlegs} use behavior cloning using human demonstrations on cheap hardware or expert demonstrations from simulation. \cite{gu2022multi, tidybot, member} use task planning \cite{strips} to chain together primitives for long-horizon tasks, particularly household object rearrangement. Yokoyama {\it et al.}~\cite{yokoyama2023asc} remove the need for a map, required by TAMP, by using expert coordination and skill correction policies. Xia {\it et al.}~\cite{xia2020relmogen} use motion planning for low-level movement and abstract the action space for reinforcement learning to end-effector space rather than joint space to achieve long-horizon goals. 
Most of these works assume simple/known object dynamics, with the notable exception of \cite{mobilepushing}.  However, in all cases, these methods require much more extensive experience than our approach.

\textbf{Imitation Learning}
Imitation learning has been applied extensively to tabletop manipulation. Chi {\it et al.}~\cite{chi2024diffusion} use a denoising diffusion inference model (DDIM) \cite{ho2020denoising} based policy. Their technique stacks a history of observations and produces a sequence of actions, allowing for multimodality and temporal continuity. Their technique learns end-to-end manipulation. Zhou {\it et al.}~\cite{zhou2023adaptiveonlinereplanningdiffusion} augment imitation learning with score-based online replanning to tackle stochastic and long-horizon tasks. Reuss {\it et al.}~\cite{reuss2023goalconditionedimitationlearningusing} show the use of imitation learning to learn goal-conditioned policies from large datasets, while Shen {\it et al.}~\cite{gailpolicy} use a non-diffusion based policy to achieve object category-level generalization.  Our work leverages imitation learning to learn a local controller from relatively few demonstrations and couples this controller to a global planner to enable zero-shot transfer to new settings.

%% file: text/4-method-new.tex
\section{Problem Formulation}

We focus on moving an attached object with difficult-to-characterize dynamics to a specified location that is far (multiple meters) from its initial location.  In particular, we concentrate on the robot-chair dynamics and chair-floor dynamics in the pulling action alone, assuming that the robot starts with a secure grasp of the object.

\textbf{Inputs}
\begin{itemize}
    \item $M$:  A known map of a room-size environment, with indicated obstacle regions, possibly gathered via one or more scans of the room with an RGB-D camera. 
    \item $x^r_t$: The robot pose within the map,  assumed to be available at all times. This can be achieved through online localization. We assume that the robot pose is known accurately (within 5 cm). 
    \item $x^o_t$: The object's pose is also assumed to be available at all times. This pose can be tracked using motion capture, estimated from point clouds, or, in our case, by tracking an April Tag \cite{apriltag} affixed to the object. 
\end{itemize}

\textbf{Output}
\begin{itemize}
    \item $a_t$: The command to the robot.  We assume that the robot has a locomotion control system that allows us to command the robot's pose specified in global coordinates.
\end{itemize}

\textbf{Demonstrations} ($\mathcal{D}$): We assume a set of long-horizon, human demonstrations of moving the attached object $\mathcal{D} = \{\tau_k\}_{k = 0}^N$, where trajectories $\tau = \{x^r_t, x^o_t\}_{t = 0}^T$. Such demonstrations can be easily collected by robot teleoperation. 

\section{Methods}

Diffusion-based behavior cloning is very good at reproducing behavior from human demonstrations. However, it tends to fail when the horizon is long or if out-of-distribution scenarios are encountered, and addressing both requires drastically scaling data. On the other hand, motion planning is good at long-horizon tasks, but struggles with planning over contact-rich tasks due to complex dynamics. To achieve generalization and robust, long-horizon reliability for mobile manipulation of unknown objects in a data-efficient manner, we propose \oursfull (\ours), combining a high-level motion planner to generate a sequence of waypoints with a low-level short-horizon diffusion policy $\pi$ to complete motion between waypoints. %

\subsection{\oursfull} \label{sec:method}
We use motion planning to provide a series of intermediate goals that $\pi$ is tasked with reaching. We assume the simple heuristic of holonomic dynamics for the robot and object and that the object's pose relative to the robot remains fixed. We first collect the environment point-cloud scan $M$ and build an offline Roadmap $R$ of object poses (see Section~\ref{experiments:motion-planning}). Given an initial $x^o_s$ and goal pose the object $x^o_g$, we ran A* algorithm \cite{Hart68} on $R$ to generate a long-horizon trajectory $\tau = \{(x^r_t, x^o_t)\}_{t=1}^T$ free of obstacle collisions. This long-horizon trajectory is downsampled by factor $f$ to generate a series of intermediate goals $g = \{(x^r_{kf}, x^o_{kf})_{k = 1}^{T/f}\}$ for the diffusion policy.

We keep a running sequence of robot and object poses in the global frame, and the short-horizon policy $\pi$ is tasked with relative movements towards the next intermediate goal, until it is sufficiently close (tested via $\proc{reached}$). We use receding horizon control, with a history length of $h_o$, a prediction horizon of $h_a$, and an execution horizon of $h_e$. 

\begin{algorithm}
    \caption{\oursfull (\ours)}
    \resizebox{0.91\textwidth}{!}{
    \begin{minipage}{\textwidth}
    \begin{algorithmic}[1]
        \Require Environment scan $M$; Initial object pose $x^o_0$, goal pose $x^o_g$; 
        
        History length $h_o$, execution horizon $h_e$.
        \State $\bm{x} = [(x^r_0, x^o_0)]$
        \State $R \gets \proc{build-roadmap}(M)$
        \State $\tau = \{(x^r_k, x^o_k)\}_{k=1}^T \gets \proc{run-motion-plan}(x^o_0, x^o_g, R)$
        \State $\bm{g} = \{(x^r_{kf}, x^o_{kf})\}_{k=1}^{T/f} \gets \proc{sample-interm-goals}(\tau)$
        \State $g \gets \bm{g}.\textit{pop}()$
        \State $t \gets 0$
        \While {not $\proc{reached}(x^o_t, x^o_g)$}
            \State $s \gets max(t - h_o + 1, 0)$
            
            \State $\bm{r_{t}}, \bm{o_{t}} = \{x^r_i\}_{i=s}^t, \{x^o_i\}_{i=s}^t \gets \proc{pad-seq}(\bm{x}, s, t)$
            
            \State $\bm{\bm{r'_{t}}, o'_{t}} \gets \proc{transform}(\left( \bm{r_{t}}, \bm{o_{t}} \right), g)$
            
            \State $\bm{a'_{t}} = \{a'_{i}\}_{i=0}^{h_a} \gets \pi(\bm{r'_t}, \bm{o'_t})$
            
            \For {$a \in \{a'_{i}\}_{i=0}^{h_e}$}
                \State $x^r_t, x^o_t \gets \proc{spot-api-execute}(a)$
                \State $\bm{x}.\textit{append}((x^r_t, x^o_t))$
                \If {$\proc{reached}(x^o_t, x^o_g)$ }
                    \State \Return{\proc{Success}}
                \EndIf
                \If {$g == (x^o_g)$ \textbf{and} $\proc{stuck}(x^r, x^o)$}
                    \State \Return{\proc{Fail}}
                \EndIf
                \If  {$\proc{lost-grasp}$}
                    \State \Return{\proc{Fail}}
                \EndIf
                \If {$\proc{reached}(x^o_t, g')$ \textbf{or} $\proc{stuck}(x^r, x^o)$}
                    \State $g \gets \bm{g}.\textit{pop}()$
                    \State \textit{break}
                \EndIf
            \EndFor
        \EndWhile
    \end{algorithmic}
    \end{minipage}
    }
    \label{alg:hierarchical}
\end{algorithm}

\begin{figure}
    \centering
    \includegraphics[width=0.9\linewidth]{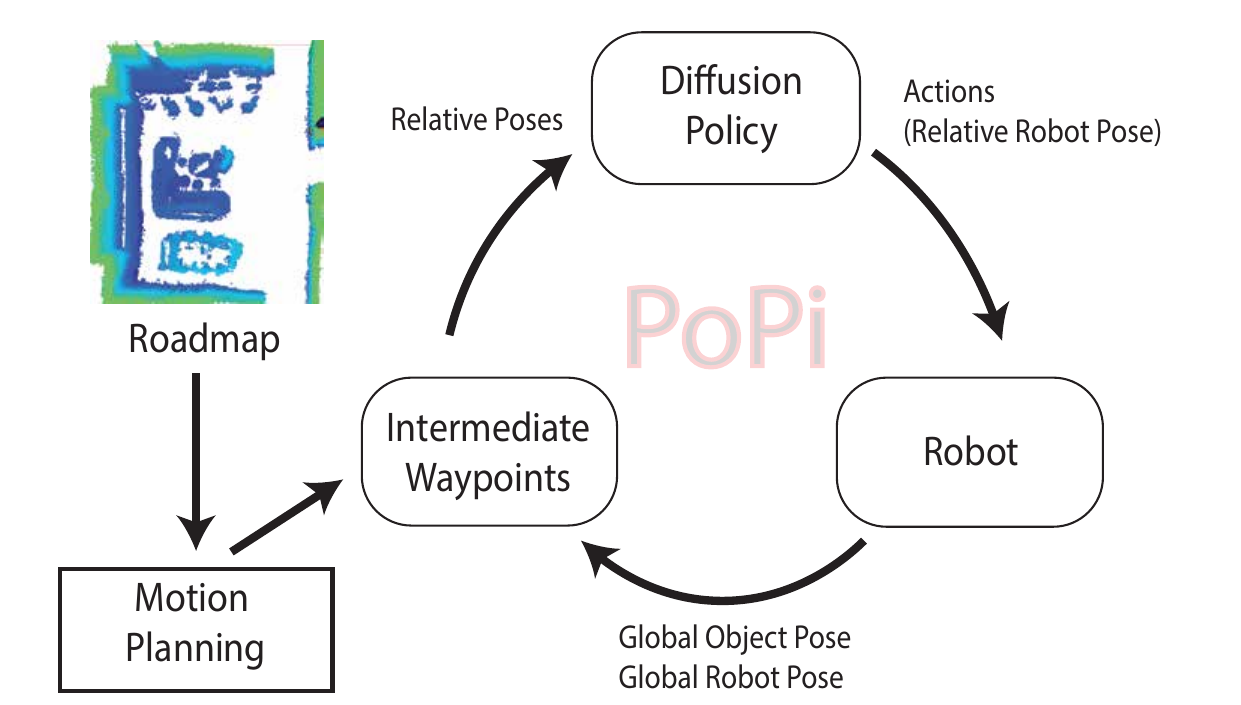}
    \caption{\oursfull}
    \label{fig:popi-algorithm}
    \vspace{-15pt}
\end{figure}

Pseudocode is depicted in \cref{alg:hierarchical} and the system is shown in \ref{fig:popi-algorithm}. The function $\proc{tranform}(poses, goal)$ returns the list of poses in the first argument  expressed relative to the goal pose in its second argument.  The function $\proc{stuck}$ detects if the robot fails to move for a pre-determined time period. 

\subsection{Short-Horizon Diffusion Policy}
\label{methods:short-horizon-training}

We use a diffusion model similar to \cite{chi2024diffusion} for conditional short-horizon action generation. The policy $\pi_\theta$ takes in a sequence of waypoint-relative object poses $\bm{o_{t}'} = \proc{Transform}(\{x^o_i\}_{i=t - h_o + 1}^t, g)$ and waypoint-relative robot poses $\bm{r_{t}'} = \proc{Transform}(\{x^r_i\}_{i=t - h_o + 1}^t, g)$, then outputs a series of waypoint-relative actions $\bm{a_{t}'} = \proc{Transform}(\{x^r_i\}_{i=t}^{t+h_a}, g)$, where actions are simply robot poses, $h_o$ and $h_a$ are history length and action horizons respectively, and $g$ is the waypoint.

The policy uses a conditional denoising network $\epsilon_\theta$ to iteratively convert random Gaussian noise $a_{t:t+h_a}^K$ into actions according to the equation 
\begin{equation*}
\begin{split}
    \bm{a'^{k-1}_t} = \alpha_k (\bm{a'^{k}_t} &- \beta_k\bm{\epsilon_\theta }(\bm{a'^{k}_t}, \bm{o_{t}'}, k, \bm{r_{t}'}) ) + \sigma_k * \mathcal{N}(0, \textbf{\textit{I}})\;\;,
\end{split}
\end{equation*}
where $\bm{a^{0}_t}$ is the denoised action sequence. We use standard noise schedule and hyperparameters $\alpha_k, \beta_k,$ and $\sigma_k$~\cite{ho2020denoising}.

To construct our training objective, we take a demonstration trajectory $\tau$. For a given point $(x^r_t, x^o_t)$ in the trajectory, we take a preceding point $(x^r_{t'}, x^o_{t'})$ with the constraint that $t > t'$ and $d(x^r_t,x^r_{t'}) < D$, where $d$ is a distance metric over robot poses and $D$ is a distance threshold that limits our policy to a short horizon. 

The first $h_o$ robot poses are taken as inputs, while the next $h_a$ robot poses are taken as actions. We build sequences $\bm{o_{t'}}$ and $\bm{r_{t'}}$, starting at $t'$ until timestep $t$ and $\bm{a_{t'}}$ starting at $t + 1$ until timestep $t+h_a$, all with padding. 

Finally, we transform these sequences relative to $x^o_g$, i.e. the goal object pose. For example, if $x^o_g \in \text{SE}(2)$, then we transform $x^r_{t'}$ with a 3x3 rigid transform $X_{x^r_{t'}}^{x^o_g}$. We do the same for object poses, robot poses, and actions. Thus, our policy sees short-horizon coordinates only and \textbf{learns to perform relative movements}. These sequences are noised with the forward process \cite{ho2020denoising} to obtain noise $\epsilon^k$ at iteration $k$. We use L1 training loss (see line 12 of \cref{alg:shorthorizontraining}).
A pseudocode description is shown in \cref{alg:shorthorizontraining}. Lines 8-9 apply head and tail padding to the data sequence to ensure that the input and output sequence stays the same length. 

\begin{algorithm}[!tp]
    \caption{Training Short-Horizon}
    \resizebox{0.91\textwidth}{!}{
    \begin{minipage}{\textwidth}
    \begin{algorithmic}[1]
        \Require Demonstration Set $\mathcal{D}$
        \Require Noise Prediction Policy $\epsilon_\theta$
        \For{$\tau \in \mathcal{D}$}
            \State $\tau \gets \{x^r_t, x^o_t\}_{t = 0}^T$
            \For{$t = 1$ \textbf{to} $T$ \textbf{and} $t' = 1$ \textbf{to} $t$}
                \If{$d(x^r_{t'}, x^r_t) < D$}
                
                    \State $g \gets x^o_t$
                    \State $s_1 \gets min(t' + h_o - 1, t)$
                    \State $s_2 \gets min(t' + h_o + h_a - 1, t)$
                    
                    \State $\bm{r_{t'}}, \bm{o_{t'}} = \{x^r_i\}_{i=t'}^{s_1}, \{x^o_i\}_{i=t'}^{s_1} \gets \proc{pad-seq}(\tau, t', s_1)$

                    \State $\bm{a_{t'}}= \{x^r_i\}_{i=s_1 + 1}^{s_2} \gets \proc{pad-seq}(\tau, s_1 + 1, s_2)$
                    \State $\bm{o'_{t'}}, \bm{r'_{t'}}, \bm{a'_{t'}} \gets \proc{Transform}\left(\left(\bm{o_{t'}}, \bm{r_{t'}}, \bm{a_{t'}}\right), g\right)$
                    \State $k \sim \text{Uniform}(1, K)$
                    \State Gradient Descent on $
                    \| \epsilon^k - \beta_k \bm{\epsilon_\theta}(\bm{a'^{k}_{t'}}, \bm{o'_{t'}}, k, \bm{r'_{t'}}) \|
                    $
                \EndIf
            \EndFor
        \EndFor
    \end{algorithmic}
    \end{minipage}
    }
    \label{alg:shorthorizontraining}
    
\end{algorithm}

%% file: text/5-experiments-new.tex
\section{Experiments}

We want to answer two questions about our method \ours:
\begin{enumerate}
    \item Can it achieve a higher long-horizon success rate in the training environment compared to baselines?
    \item Can it generalize to environments, objects, and grasp poses that are different from those in training?
\end{enumerate}

\subsection{Task and Metrics}

The task requires manipulating a five-wheeled office chair into a goal pose in the presence of obstacles. These chairs have many internal degrees of freedom. Notably, the wheels on the legs rotate passively, and the friction between the wheels and the ground is difficult to accurately model and simulate, especially on carpet. The training environment and robot are depicted in \cref{fig:spot-chair} and a birds-eye view is shown in \cref{fig:mapping:training-distribution}. We focus on the movement only, assuming that the grasping of the chair has already been done and that the grasping point is near the center top on the back of the chair. 

Given that the policies are trained on trajectories whose length range from 8 m to 18 m, we test goals sampled from 2 m, 6 m, and 10 m away. Correspondingly, it takes 1, 2, and 3 turns to achieve those goals. We measure task success rate, where an evaluation is deemed successful if the chair reached within 30 cm of the target position.

As we are assuming minimal knowledge of the object/environment dynamics, we are particularly interested in how our method performs in situations that differ from training (unseen environment with carpeted floor, unseen chair, different grasp pose). This gives us in total eight conditions to test all methods, only one of which is in-distribution. The unseen environment is depicted in the bottom of \cref{fig:spot-chair}, while the unseen chair and grasp poses are shown in \cref{fig:generalization}.

\subsection{Hardware Setup}
\begin{figure}[!tp]
    \centering
        \includegraphics[width=0.8\linewidth]{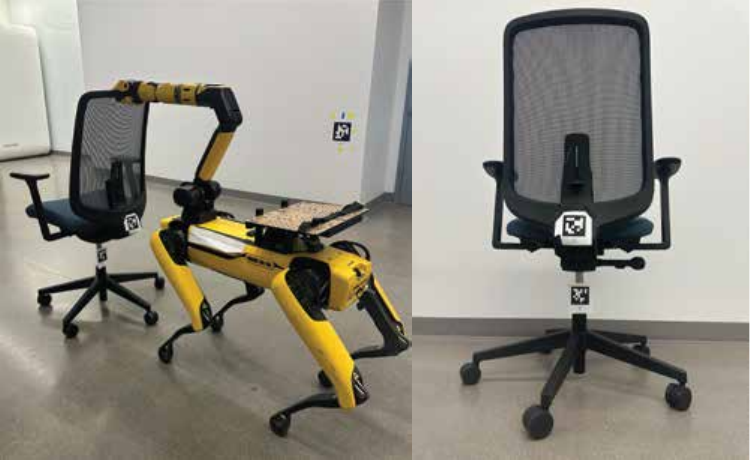}
        \caption{Experimental Setup. Left is the robot and chair. An AprilTag for localization is also shown in the background. The right shows the AprilTag setup affixed to the chair.}
        \label{fig:setup}
        \vspace{-15pt}

\end{figure}

We use the Boston Dynamics Spot robot (a large quadruped) as the mobile manipulator. The robot is equipped with a 6-dof arm with a simple claw gripper to grasp the back of a chair. The robot has access to six cameras (five when manipulating) each with RGB-D information that it uses for odometry. We command SE(2) pose of the robot, and low-level control is done by its official black-box API. Cameras on the front of the robot read an April tag on the chair to observe the chair's SE(2) pose as shown in \cref{fig:setup}. All global poses are computed relative to a fixed fiducial in the environment. 

\subsection{Motion Planning}
\label{experiments:motion-planning}
To generate a 2D road-map for planning, we represent the robot with a rectangle of size 1.1 m x 0.5 m and the chair forms a circle of radius 0.3 m, with the centers separated by 0.7 m (as illustrated in \cref{fig:long-horizon-testing}). We take a grid of points in SE(2) space evenly spaced 10 cm/10\degree apart (corresponding to the chair pose), filter out those where the chair or robot are in collision with obstacles, and connect adjacent points. This forms a road-map for motion planning. 

Here, we assume that the robot-chair pose is fixed, that the robot is directly behind the chair, and that the robot-chair system can move holonomically. In other words, the system is rigid and can move incrementally in any direction plus rotate incrementally either clockwise or counterclockwise. This model is quite impoverished; it does not take into account the intricacies of displacement and force necessary to move the chair in a given direction, e.g. if the wheels are oriented perpendicularly to the desired motion.

\begin{figure}[]
    \centering
    \includegraphics[width=0.74\linewidth]{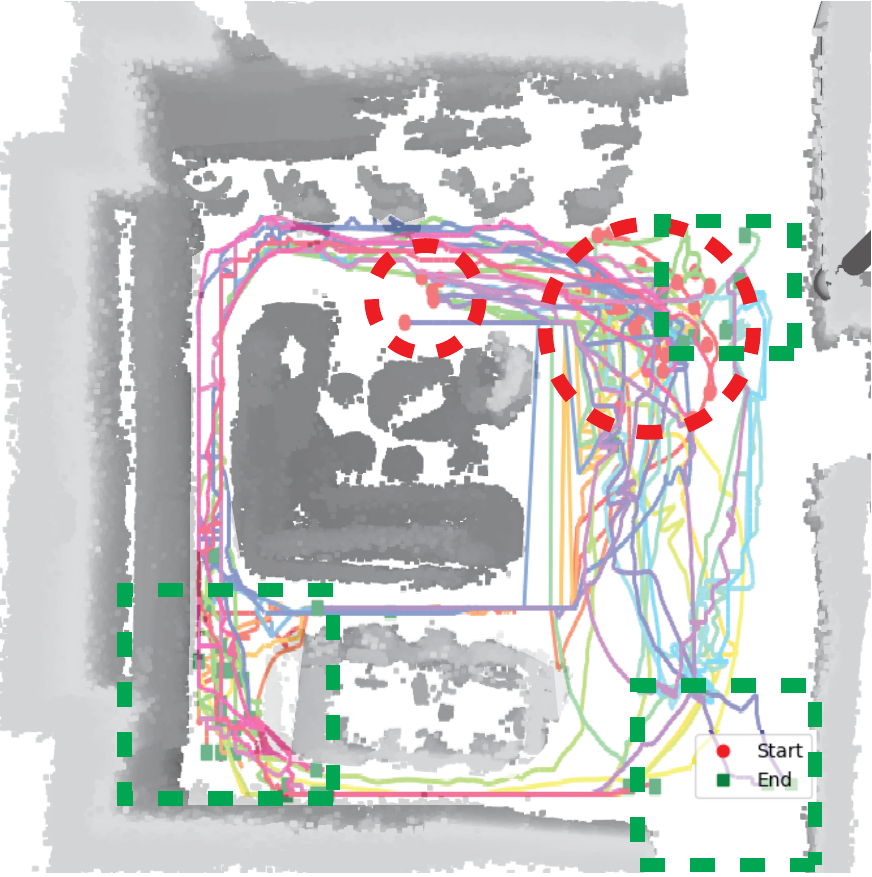}
    \caption{Map of the training environment (in grayscale) with demonstration trajectories overlaid. Starting points are shown as red circles, roughly drawn from the respective dashed regions. Endpoints are shown as green squares, roughly drawn from the respective dashed regions.}
    \label{fig:mapping:training-distribution}
    \vspace{-15pt}
    
\end{figure}

\begin{table*}[ht]
    \centering
    \begin{tabularx}{\textwidth}{X X X X X X X}
        \toprule
        \textbf{Goal Distance} & \textbf{Rotation} & \textbf{PoPi} & \textbf{RRT} & \textbf{A*} & \textbf{Local Diffusion} & \textbf{Global Diffusion} \\
        \midrule
        \textbf{10 m} & \textbf{270°} & \textbf{8/10} & 5/10 & 2/10 & 0/10 & 0/10 \\
        \textbf{6 m} & \textbf{180°} & \textbf{8/10} & 6/10 & 2/10 & 2/10 & 0/10\\
        \textbf{2 m} & \textbf{90°} & \textbf{10/10} & \textbf{10/10} & 3/10 & 8/10 & 0/10\\
        \bottomrule
    \end{tabularx}
    \caption{The number of successful trials out of all trials at increasing distance between the chair's goal and starting locations and with progressively more turns around obstacles (1, 2, 3). \ours significantly outperforms the other baselines.}
    \label{tab:long-horizon}
    \vspace{-10pt}
\end{table*}

\subsection{Baselines and Ablations}
We consider two baselines:  pure diffusion policies and pure motion planning.

\textbf{Pure Diffusion} We use the same demonstrations $\mathcal{D}$ to train a long-horizon diffusion policy in the global frame, similar to \cite{janner2022diffuser}. In \cref{alg:shorthorizontraining}, we simply remove lines 3-4 and 8-9 and fix $t = T$, i.e. the goal is fixed to the final position in the trajectory, and there is no restriction on the preceding action/observation sequences. However, we note that compared to the method in that paper, we have significantly fewer demonstrations.
At evaluation time, {\em there is no planner} and we constrain trajectories with the same methods from \cite{janner2022diffuser}, i.e. preventing them from passing through obstacles. Thus, the mapping information is explicitly used by the policy, in addition to implicit information about obstacles in the training data. We apply the same receding horizon control as a comparison.
As an additional diffusion baseline, we also compare our "local" short-horizon diffusion policy with no motion planning.

\textbf{Pure planning} As a second baseline, we apply both shortest path search using A* in the roadmap and an online RRT to navigate between intermediate waypoints of the global trajectory. We simply replace line 11 in \cref{alg:hierarchical} (where we call $\pi$) with a call to the respective planner. Furthermore, the planners do not use a history, only the current pose. The online RRT method takes into account the current chair-pose relative to the robot, which may change over time, while shortest path search does not. However, both assume that straight line, holonomic movements are possible and ignore the dynamics of the chair.

\subsection{Training}
We collected 35 demonstrations in the environment shown in the top of \cref{fig:spot-chair} starting and ending at various places to cover all movement within the environment. \cref{fig:mapping:training-distribution} shows the trajectories used to train both \ours and a global diffusion policy. The global diffusion policy is trained on whole trajectories and thus has 35 examples. The diffusion submodule of \ours{} is trained on relative snippets, which allows substantial data reuse giving 36,000 examples. 

\subsection{Long-horizon performance}

We begin by studying these methods in long-horizon tasks with the same conditions (i.e. floor, grasp, chair) as training.

\begin{figure}[]
    \centering
    \includegraphics[width=0.74\linewidth]{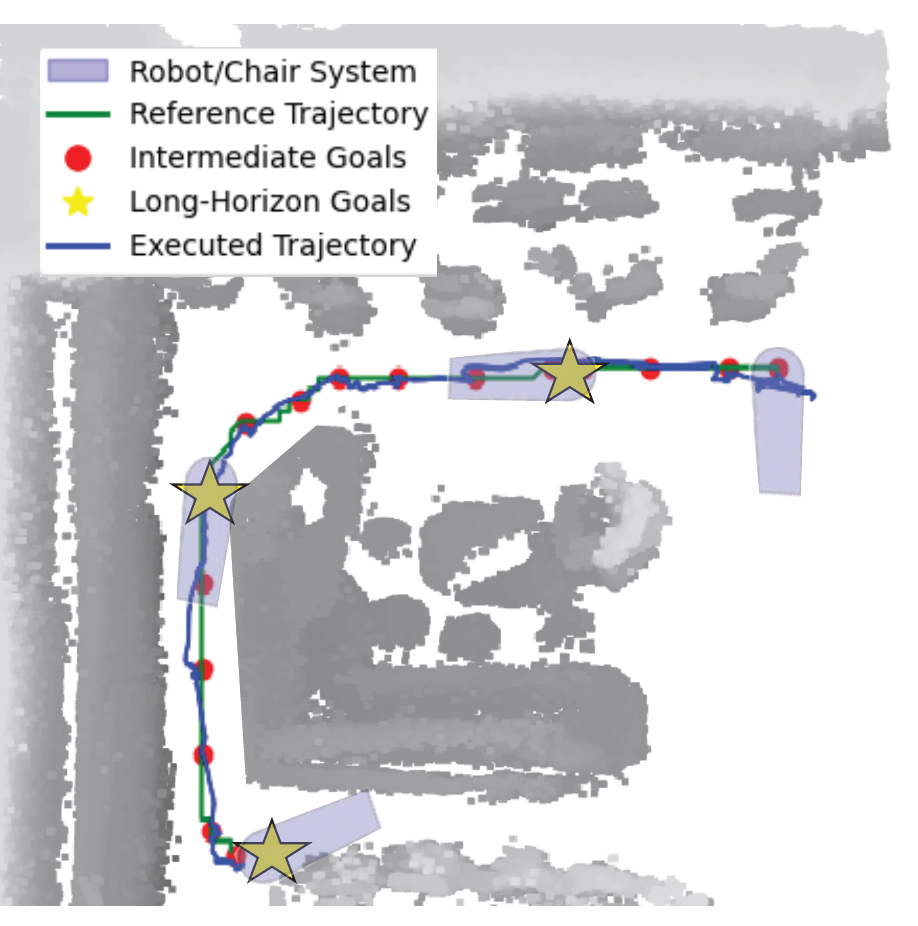}
    \caption{Trajectory to test long-horizon success. Three long-horizon goals are given at 2 m, 6 m, and 10 m. An example execution of \ours is shown here in blue. The light blue shape corresponds to the robot/chair system as described in \cref{experiments:motion-planning}}
    \label{fig:long-horizon-testing}
    \vspace{-15pt}
    
\end{figure}

We report the success rate, where an evaluation is deemed successful if the chair reached within 30 cm of the target position. We tested each method in the training environment with trajectories of varying horizon and curvature with goals at 2 m, 6 m, and 10 m distance requiring 1, 2, and 3 turns to get around obstacles. For testing, we placed an extra obstacle in the center blocking off the narrow passageway so that the robot must take the longer 10 m route with more turns to reach its goal. At 2 m, there is minimal obstacle interaction, while at 10 m, the robot must avoid obstacles for almost half of the trajectory. The 10 m testing trajectory is depicted in \cref{fig:long-horizon-testing} in the same environment as \cref{fig:setup} along with an example execution (using \ours). Results are shown in \cref{tab:long-horizon}. 

\begin{table*}[ht]
    \centering
    \begin{tabularx}{\textwidth}{X X X X X X}
        \toprule
        \textbf{Environment} & \textbf{Chair Type} & \textbf{Grasp Type} & \textbf{PoPi} & \textbf{RRT} & \textbf{A*} \\ 
        \midrule
        
        \multirow{4}{*}{Training Environment} 
        & Training Chair  & Training Grasp  & \textbf{8/10} & 5/10 & 2/10  \\ 
        & Training Chair  & Unseen Grasp    & \textbf{5/10} & 3/10 & 3/10   \\ 
        & Unseen Chair    & Training Grasp  & \textbf{2/10} & 0/10 & 1/10   \\ 
        & Unseen Chair    & Unseen Grasp    & \textbf{1/10} & 0/10 & \textbf{1/10}   \\ 
        \midrule
        
        \multirow{4}{*}{Unseen Environment} 
        & Training Chair  & Training Grasp  & \textbf{7/10} & 3/10 & 3/10  \\ 
        & Training Chair  & Unseen Grasp    & \textbf{6/10} & 5/10 & 5/10  \\ 
        & Unseen Chair    & Training Grasp  & \textbf{1/10} & 0/10 & 0/10  \\ 
        & Unseen Chair    & Unseen Grasp    & \textbf{5/10} & 1/10 & 3/10   \\ 
        \bottomrule
    \end{tabularx}
    \caption{Success rate across a variety of unseen conditions.}
    \label{tab:generalization}
    \vspace{-10pt}
    
\end{table*}

We find that as the horizon increases, the performance of each of the methods decreases. The primary failure modes are (a) losing grip of the chair due to difficult dynamics and (b) collisions with obstacles that prevent chair movement. At the longest horizon, \ours performs the best, achieving 80\% success compared to 50\% for the next best baseline (RRT). The baseline using $A*$ only achieves 30\% success at the short-horizon, and 20\% at medium and long-horizon, with the remaining trials failing  by losing grasp very quickly. RRT does better, presumably because it incorporates the current relative pose of the chair in the robot frame, which may be different than the rigid pose assumed by $A*$. However, it still fails to achieve long-horizon robustness, as the simple dynamics model leads to failure half of the time. 

We find that the global diffusion baseline is unable to achieve any success. The trajectories generated are sensible, however, failure by lost grasp occurs almost immediately. \cite{janner2022diffuser} shows that with substantial amounts of demonstrations in constrained distributions, this method generates good trajectories. However, the global diffusion policy is unable to learn the dynamics that enable long-horizon goals in our more challenging setting, presumably because of limited demonstration data. The pure short-horizon diffusion baseline works very well at 2 m (80\% success). As the training method from \cref{methods:short-horizon-training} takes many short snippets per demonstration, the effective amount of training data is much higher (36,000 snippets), and this leads to robust performance. However, without the motion planning to avoid obstacles, it is unable to perform well beyond short horizons.

\subsection{\oursfull Generalizes Better}
We chose a separate testing environment to evaluate the generalization of our methods to different obstacle configurations. The chosen environment has additional dynamics due to high friction from the carpeted floor that were unseen in the training environment. The environment and testing trajectory is depicted in the bottom of \cref{fig:spot-chair}. To compare across environments, we choose a trajectory with 10 m displacement and 2 turns in both environments. 

\begin{figure}[]
    \centering
    \includegraphics[width=0.8\linewidth]{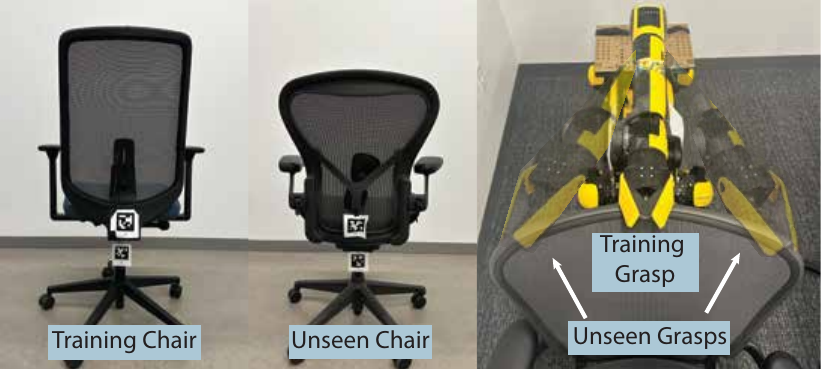}
    \caption{Depiction of the variables used to test generalization}
    \vspace{-20pt}
    \label{fig:generalization}
    
\end{figure}

To test generalizability across objects, we tested manipulation using a different chair and also varied the initial grasp pose to test robustness to the obstacle's initial position. Both variations are depicted in  \cref{fig:generalization}. Results are shown in \cref{tab:generalization}. 

We did not attempt the global diffusion baseline in the new environment. The map of the training environment is implicitly encoded in the training distribution, and therefore, we do not expect it to perform with any success. 

We find that the other methods retain some performance as the obstacles change, reflecting that motion planning is robust to changes in obstacle arrangement. We find that \ours generalizes much better than the pure motion planning baselines across environments, with success rate staying high (70\%) even in the new environment. As we change to the unseen grasp in the training environment with the training chair, \ours maintains 50\% success rate, while the baselines achieve only 30\%.  Interestingly, we note that all of the methods, when using the unseen grasp, perform better in the new environment than the training environment.  We conjecture that in the new environment, the carpet's additional friction reduces acceleration of the chair, which is a major cause of lost grasps. We note that for the unseen chair + training grasp, all methods fail by losing grasp, with rare success. The design of the unseen chair made grasping the center more unstable. The unseen grasp, although it did not help in the training environment (with lower friction floor and higher acceleration) is much more robust in the new environment (with higher friction and lower acceleration). We see that \ours achieves 50\% success with the unseen grasp + unseen chair in the new environment compared to 10\% and 30\% by the motion planning baselines. Furthermore, \ours consistently outperforms the baselines in all eight scenarios.

%% file: text/6-conclusion.tex
\section{Conclusion}

In this paper we describe \oursfull, a hierarchical algorithm for long-horizon robot manipulation problems where world dynamics are partially unknown.  We find that \ours performs consistently
better as the horizon increases, compared to a "global" diffusion policy or motion
planning assuming a rigidly-attached object. Importantly, \ours generalizes to new layouts, grasps, chairs, and
even flooring, without any further training.

One obvious limitation of \ours is the inability to recover from complete failure, so incorporating both manipulation and grasping would improve its success rate. Future work includes incorporating point-cloud observations and extending this framework to other loco-manipulation tasks.

%% file: text/7-acknowledgements.tex
\section{Acknowledgements   }
We gratefully acknowledge support from AI Singapore AISG2-RP-2020-016; from NSF grant 2214177; from AFOSR grant FA9550-22-1-0249; from ONR MURI grant N00014-22-1-2740; from ARO grant W911NF-23-1-0034; from the MIT Quest for Intelligence; and from the Artificial Intelligence Institute.